\documentclass[mlmain,twocolumn]{jmlr}

\usepackage[hpos=300px,vpos=70px]{draftwatermark}
\usepackage{booktabs}
\SetWatermarkText{\test}
\SetWatermarkScale{1}
\SetWatermarkAngle{0}
\usepackage{longtable}% for long tables

\usepackage{booktabs}
\usepackage[load-configurations=version-1]{siunitx} % newer version
\theorembodyfont{\upshape}
\theoremheaderfont{\scshape}
\theorempostheader{:}
\theoremsep{\newline}

\usepackage{amsmath,amsfonts,bm}

\def\eqref#1{equation~\ref{#1}}
\def\1{\bm{1}}

\DeclareMathAlphabet{\mathsfit}{\encodingdefault}{\sfdefault}{m}{sl}
\SetMathAlphabet{\mathsfit}{bold}{\encodingdefault}{\sfdefault}{bx}{n}

\jmlrvolume{}
\firstpageno{1}
\jmlryear{2022}
\jmlrworkshop{}

\title[PainPoints]{PainPoints: A Framework for Language-based Detection of Chronic Pain and Expert-Collaborative Text-Summarization}
  \author{\Name{Shreyas Fadnavis$^{\dagger}$} \Email{sfadnavis@mgh.harvard.edu}\\
  \Name{Amit Dhurandhar$^*$} \Email{adhuran@us.ibm.com}\\
  \Name{Raquel Norel$^*$} \Email{rnorel@us.ibm.com}\\
  \Name{Jenna Reinen$^*$} \Email{jenna.reinen@ibm.com}\\
  \Name{Carla Agurto$^*$} \Email{carla.agurto@ibm.com}\\
  \Name{Erica Secchettin$^{\ddagger}$} \Email{erica.secchettin@aovr.veneto.it}\\
  \Name{Vittorio Schweiger$^{\ddagger}$} \Email{vittorio.schweiger@univr.it}\\
  \Name{Giovanni Perini$^{\ddagger}$} \Email{giovanniperini1@gmail.com}\\
  \Name{Guillermo Cecchi$^*$} \Email{gcecchi@us.ibm.com}\\
  \addr $^{\dagger}$Harvard Medical School, $^*$IBM Research, $^{\ddagger}$University of Verona}

\begin{document}
\maketitle
\begin{abstract}
Chronic pain is a pervasive disorder which is often very disabling and is associated with comorbidities such as depression and anxiety. Neuropathic Pain (NP) is a common sub-type which is often caused due to nerve damage and has a known pathophysiology. Another common sub-type is Fibromyalgia (FM) which is described as musculoskeletal, diffuse pain that is widespread through the body. The pathophysiology of FM is poorly understood, making it very hard to diagnose. Standard medications and treatments for FM and NP differ from one another and if misdiagnosed it can exacerbate symptoms and lead to poor outcomes. To overcome this difficulty, we leveraged the known differences in these pain types to propose a novel framework, PainPoints, which accurately detects the sub-type of pain and generates clinical notes via summarizing the patient interviews. Specifically, PainPoints makes use of large language models to perform sentence-level classification of the text obtained from interviews of FM and NP patients with a reliable AUC of 0.83. Using a sufficiency based interpretability approach, we explain how the fine-tuned model accurately picks up on the nuances that patients use to describe their pain. Finally, we generate summaries of these interviews via expert interventions by introducing a novel facet-based approach. PainPoints thus enables practitioners to add/drop facets and generate a custom summary based on the notion of ``facet-coverage" which is also introduced in this work.
\end{abstract}
\begin{figure*}[ht]
\centering
\includegraphics[width=1\textwidth]{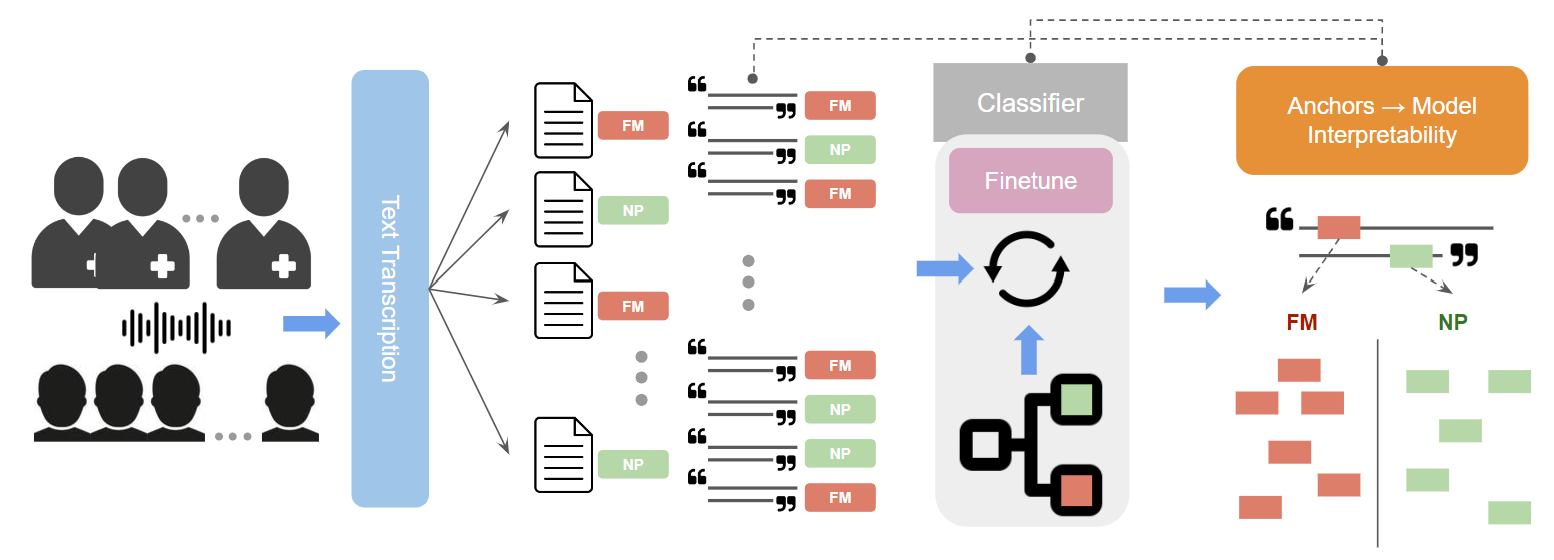}
\caption{Depicts the flow of PainPoints, where the recordings of the interviews are transcribed into text documents for each patient. Individual sentences are extracted and fed into the classifier. After fine-tuning the classifier, anchor words are extracted from each class (FM and NP) for downstream analysis.}
\label{fig:flow_1}
\end{figure*}

\section{Introduction}
\label{sec:intro}
Chronic pain is a disorder that can severely impact function and is characterized by a complex pathophysiology. A recent NIH study reported that it affects approximately 20.4\% of the population in the United States, from which approximately 7\% experience high-impact pain \citep{dahlhamer2018prevalence, zelaya2020chronic}. Not only does chronic pain impact a substantial number of people but it also leads to heavy healthcare costs for medication, treatment, pain management and care. In this work we focus on two commonly occurring sub-types of chronic pain: \textbf{1) Neuropathic Pain (NP)} and \textbf{2) Fibromyalgia (FM)}. NP is typically caused by injuries damaging the peripheral nerves or the spinal cord. Damage to the central nervous system disrupts the flow of neural information resulting in the sensation of pain \citep{colloca2017neuropathic}. This often causes inflammation or swelling and it is easier to diagnose. FM is a more widespread and diffuse type of pain which is typically characterized by muscle tenderness \citep{hauser2015fibromyalgia}. This type of pain is multi-site, i.e., diffused throughout the body, and is not associated with inflammation like NP. This type of pain is more difficult to diagnose and seldom exhibits a specific location of manifestation.

Chronic pain is diagnosed via clinical interviews in which a clinician examines the patient using a particular set of questions. Based on patient response, the pain is scored and the sub-type is determined. However, this process is noisy owing to many factors such as differences in manifestation, associated disorders, and subjectivity and experience of the interviewer. \citep{breivik2008assessment}. Moreover, misdiagnosis of the type of pain can change the patient's treatment trajectory dramatically, leading to ineffective treatment and prolonged symptom severity. Here, we leveraged the subtle nuances in particular sub-types of chronic pain which may manifest in linguistic features. For instance, NP is often described as shooting, stabbing, burning or shock like sensations. While a primary cause of NP is injury, it can also be caused due to multiple sclerosis \citep{ehde2005chronic}, transverse myelitis \citep{frohman2010transverse} and stroke \citep{dahlhamer2018prevalence}. NP is very disabling often causing loss of sleep, reduced socializing and reduced interest in daily activities in life. While easily diagnosed, NP is less responsive to treatments.

Fibromyalgia, on the other hand, is characterized by musculoskeletal pain. It has an unknown etiology making it harder to detect. A common theory is that FM, due its multi-site widespread manifestation, affects the way pain is regulated by the central nervous system. FM is known to cause muscle tenderness and increased sensitivity \citep{buskila1993assessment}. Some key symptoms include stiffness, fatigue and mind-fog. In addition, its diagnosis depends on history of the particular patient and physical exams. It is also commonly associated with co-morbidities such as depression, anxiety, sleep disorders, and concentration issues. Apart from these, FM patients experience higher rates of irritable bowel syndrome (IBS) \citep{sperber1999fibromyalgia}, dry eyes \citep{bonafede1995association}, gastroeshophaegal reflux disease (heartburn) \citep{wang2017bidirectional} and orthostatic hypotension (lightheadedness) \citep{martinez2007biology}. While these symptoms commonly occur during diagnosis, they by no means cover the entire spectrum of symptoms and associated diseases. The manifestation of FM chronic pain varies per patient and is associated with distinct social/ emotional aspects, as well as demographic differences in culture and gender. 

In this work, we hypothesized that linguistic features manifesting from the distinct pain characteristics of NP and FM pain would allow us to develop a novel framework using clinical interviews to distinguish between them \citep{berger2021, berger2022}. We demonstrated that PainPoints 1) accurately detects the mentioned sub-types of pain FM and NP by formulating a sentence-level classification problem, 2) generates summaries of the interviews using interpretable machine learning, and 3) allows for expert-in-the-loop learning such that the medical practitioner can intervene on the summary and re-summarize the interview by looking at the `facets' present. 
\begin{figure*}[ht]
\centering
\includegraphics[width=1\textwidth]{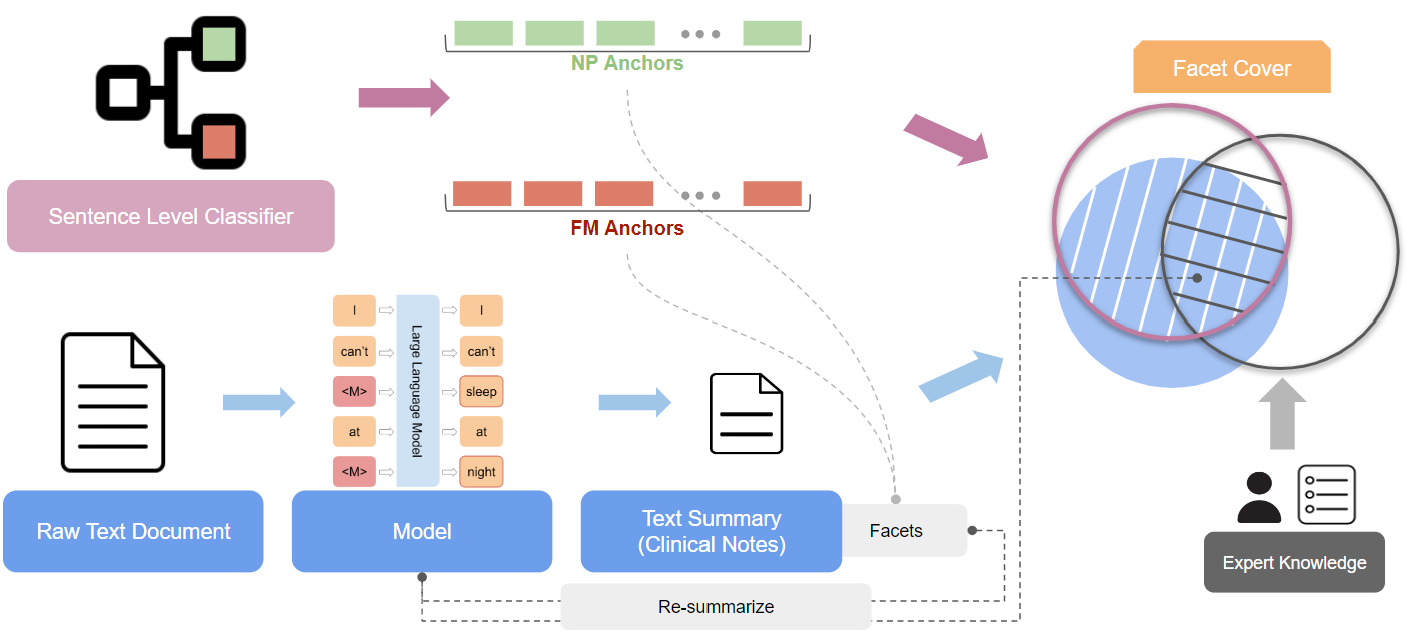}
\caption{Delineates the notion of how the overlap between the anchors extracted from the predictive model and the text summary is computed. Optionally, expert facets could also be included to calibrate and re-summarize the text.}
\label{fig:flow_2}
\end{figure*}
\looseness=-1
\section{Background and Related Work}
\looseness=-1
\textbf{Large Language Models:} Language models have undergone a tremendous revolution since 2017 due to the adoption of attention-based modeling and introduction of transformers \citep{vaswani2017attention}. Rather than extracting intelligent features from the data and performing the learning task on them, current deep learning models are set up using a self-supervised task. Simply put, the models learn by masking out a part of the sentence and predicting the masked out part using some context. This is a paradigm shift different from traditional NLP models where semantic information was used to gain a deeper understanding of the context and design better features. In the newer self-supervised models, the training is on large amounts of data with a huge number of parameters. The first successful models such ELMo \citep{peters2018dissecting} and ULMFiT \citep{howard2018universal} used LSTMs as the backbone to learn word representations and showed improvements over previous approaches such as GloVe \citep{pennington2014glove}. Later it was shown that transformer-based models achieved better performance on a variety of tasks in GPT (800M words) \citep{radford2018improving}, BERT (3.3B words) \citep{devlin2018bert} and GPT2 (40B words) \citep{radford2019language}. Over the past two years, several more models yielding better performance have been proposed. These include RoBERTa \citep{liu2019roberta}, XLNET \citep{yang2019xlnet}, ERNIE \citep{sun2019ernie}, GPT-3 \citep{brown2020language}, etc. However, the performance gain obtained from increasing the size of the model is not as much. Understanding the scaling laws of these large language models (LLMs) has emerged as an active field of research where optimal and compute-efficient strategies of training these models are explored \citep{kaplan2020scaling}.

A key outcome of the revolution of these LLMs is the modularity with which these pre-trained models can be used. Not only does this allow for editing and fine-tunining the model on custom data, but it also opens up new opportunities for building human-collaborative AI systems \citep{lee2022coauthor}. While PainPoints can work with any LLM, in this work, we use BERT given its widespread adoption. Apart from being a deeply bi-directional model, its architecture allows for parallel training. Along with a novel masking approach, BERT also breaks each word into sub-words which are used as tokens for model training. 

\looseness=-1
Pre-trained LLMs can be used in two ways: 1) Feature based -- where the embeddings from the model can be directly used to perform the task on new data, and 2) Fine-tuning based: where the model is fine-tuned on the dataset on which the learning task is performed. Specifically for BERT, the fine-tuning is done by using the masked language modeling (MLM) approach. BERT also uses a self-attention mechanism to perform the fine-tuning and is relatively fast and compute-efficient. For our chronic pain cohort, we follow the same procedure described in \citep{devlin2018bert} to fine-tune on the sentence-classification task. \\
\textbf{Interpreting Models and Explaining Predictions:} Model interpretability, also known as explainable AI, is a field of machine learning which aims to understand and explain predictions made from `black-box' models. As mentioned above, LLMs are progressively increasing in the number of parameters and the amount of data it is trained on. For example, BERT is trained on 3.3B words and has 110M parameters. A common approach to explain such LLMs is by performing a post hoc interpretability analysis. Local interpretable model-agnostic explanations (LIME) \citep{ribeiro2016should} and SHapley Additive exPlanations (SHAP) \citep{lundberg2017unified} are the most common interpretability approaches that are widely adopted for a variety of data. These approaches approximate the complex non-linear decision boundary of the LLM by a linear model fit locally in the post hoc step. LIME also enforces the locally linear model to be sparse so that it is easier to interpret the large number of data points. A common step in these models is to perturb the inputs given to the model to gauge which features hamper the predictive performance of the predictions the most. LIME and SHAP allow the user to locally understand the relative importance of the features by representing them as a linear combination. However, it is unclear to the user what the explanation would look like on an unseen test sample, since they do not reveal any global information and can have the similar features exhibiting different importance levels along the decision boundary. While tools like ProfWeight \citep{dhurandhar2018improving}, SRatio \citep{dhurandhar2020enhancing} and Knowledge Distillation \citep{hinton2015distilling} can be used to get a global understanding of the model, the fine-grained understanding of the features and data points is lost. 
\looseness=-1
To address this issue, the Anchors method was proposed \citep{ribeiro2018anchors}, which can be seen as an extension the LIME. Specifically, Anchors finds features such that even after perturbations, the role of the feature does not change. For example in our case of text features, anchors would yield words in the sentence which remain unchanged even when the remaining sentence is perturbed. These `anchor' words sufficiently explain the prediction made by the model by being able to recover the original sentences prediction. Thus, it serves as a useful tool for us, as it reveals exactly which words in the sentence led to predicted class. Anchors are also more intuitive for the setting of human-AI collaborative systems \citep{lee2022coauthor} as they give definitive words/features which are easy to understand and useful in the downstream tasks. In a Sec.~3 we detail how anchors can be used to re-summarize clinical interviews with expert-in-the-loop intervention. Other methods such as input reduction \citep{wallace2019allennlp} and path-sufficient explanations \citep{luss2021towards} have been proposed. While we use anchors for our experiments, our approach is still applicable to these interpretability methods. \\
\looseness=-1
\textbf{Expert-Collaborative Modeling:} In this work, we focus on the task of text summarization to generate clinical notes from clinical interviews. To do so, we want experts (medical practitioners) to intervene on the model generated summaries. LLMs have shown unprecedented capabilities of generating text, however, they are very context-driven. The summary highly depends on the data the LLMs have been trained on. Even if fine-tuned on a specific dataset, there is a chance that the LLMs can miss out on key facets due to missing the context or the size of the summary. This hampers its direct use as a tool for clinical summarization. It is also difficult to check the summary's quality as the interpretation of a `good summary' is very subjective. Metrics like ROUGE \citep{lin2004rouge} and BLEU \citep{papineni2002bleu} measure the precision and recall of the summary using the overlap of words. This is however not very useful in our case as these scores increase in proportion with the size of the summary, not revealing if important ideas were captured in the summary. To address this issue, we introduce the idea of \textit{Facet Coverage} as a new measure to quality check the summary. We also detail how a medical practitioner can interact with the facets present or missing in the summary to re-summarize the interview.

% The basic idea is to use the facets, i.e. anchors from the model predictions of the sentence-level classifier to check if the text summary missed key nuances or important points from the original interview document.

\begin{table}[]
\centering
\begin{tabular}{@{}ll@{}}
\toprule
Model                                         & AUC                       \\ \midrule
\multicolumn{1}{|l|}{BERT + SVM}              & \multicolumn{1}{l|}{0.62} \\ \midrule
\multicolumn{1}{|l|}{BERT + Gaussian Process} & \multicolumn{1}{l|}{0.59} \\ \midrule
\multicolumn{1}{|l|}{BERT + MLP}              & \multicolumn{1}{l|}{0.59} \\ \midrule
\multicolumn{1}{|l|}{BERT + AdaBoost}         & \multicolumn{1}{l|}{0.59} \\ \midrule
\multicolumn{1}{|l|}{Fine-tuned BERT}         & \multicolumn{1}{l|}{0.83} \\ \bottomrule
\end{tabular}
\caption{Comparison of mean AUC scores.}
\label{tab:auc_table}
\end{table}
\begin{figure*}[ht]
\centering
\includegraphics[width=1\textwidth]{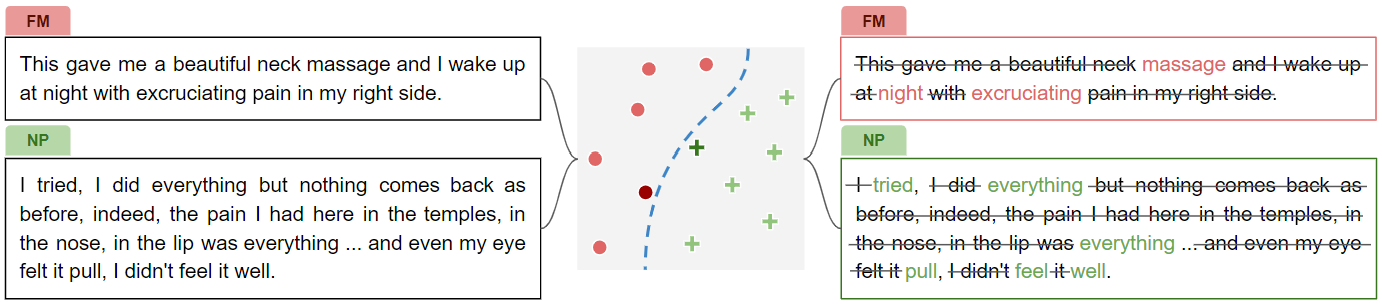}
\caption{Anchor words picked up by the fine-tuned BERT model on two randomly chosen sentences from the FM and NP cohort each.}
\label{fig:ss_anchors}
\end{figure*}
\section{Methods}
In this section we explain the implementation of each component of PainPoints:\\
\textbf{Sentence Level Classification:} Text classification is typically done at a document level, sentence level or sub-sentence (phrase) level. In PainPoints we chose to train a sentence-level classifier because we want to interpret the subtle nuances in the text that affect the binary classification task. If the entire document was used, the information extracted would be coarse-grained and lost in the process. In clinical analyses, the size of the cohorts is typically very small, therefore performing a document-level classification is not possible. While phrase-level classification can be interesting in this analysis, it is unclear how to extract these phrases. 

We made use of bidirectional encoder representations from transformers (BERT) model to perform this sentence classification. We started by using the pre-trained BERT model to perform the sentence-level classification. The embeddings generated from the pre-trained BERT model were used as features to different classifiers. In this study we compared SVM \citep{cortes1995support, pisner2020support}, Gaussian process \citep{nickisch2008approximations}, MLP-Neural Network \citep{lecun2015deep} and AdaBoost \citep{schapire2013explaining} as classifiers on top of the BERT embeddings. Next, we fine-tuned the BERT model on our data and then performed the sentence-level classification. Our dataset consisted of a total of 55 documents (40 FM and 15 NP) generated from transcribing the interviews \citep{scandola2021bodily}. The FM group recruitment method and characteristics are described in \cite{schweiger2022quality}. A smaller sample of patients (N=15) suffering exclusively from neuropathic pain (NP group) has been specifically recruited for the purposes of this study as a comparison with the FM group. NP patients were recruited with the same method and in the same pain therapy center as FM patients, and signed a formal consent to participate and were recorded with the same procedure. Each document was split into sentences using NLTK \citep{loper2002nltk} and each sentence in the document was given the label associated with it (FM or NP), resulting in 5970 sentences. We made use of the BERT tokenizer to get word tokens from each sentence. This is required owing to the MLM technique that BERT employs internally \citep{devlin2018bert}. [CLS] and [SEP] tokens were added to mark the beginning and ending of each sentence with a maximum length of 64. As per recommendation, the shorter sentences were padded on the right and attention masks were set to 0/1 accordingly. The learning rate was set to $1\mathrm{e}{-5}$ and the model was fine-tuned for 2 epochs. To extract the embeddings from the pre-trained model, values of the hidden units in the last 4 layers were averaged. To fine-tune the BERT model, sentences were shuffled and split into 75\% training, 15\% validation and 10\% test sets. We make use of the same train-validate-test ratios of both the pre-trained and fine-tuned models. 
\begin{figure}[ht]
\centering
\includegraphics[width=.45\textwidth]{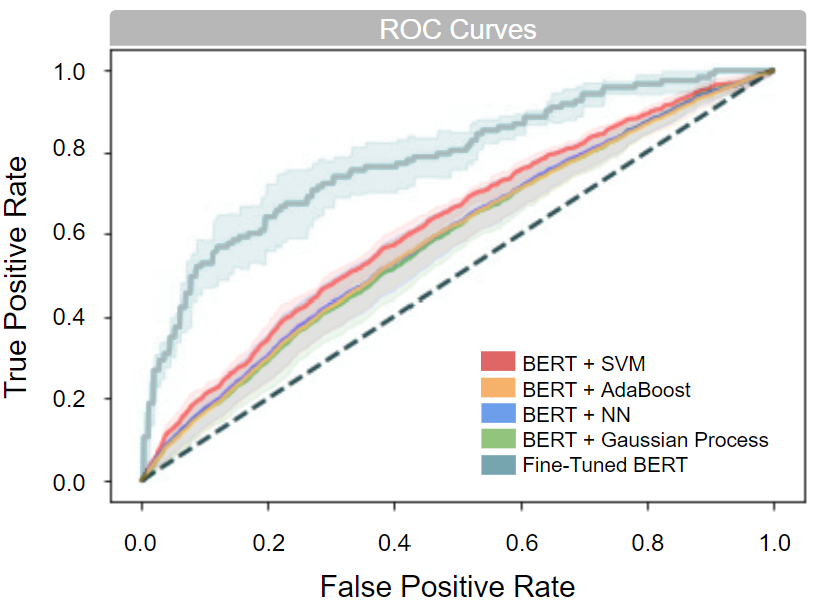}
\caption{ROC curves corresponding to the different classifiers.}
\label{fig:rocs}
\end{figure}\\
\looseness=-1
\textbf{Interpretability via Anchors:} Given that the LLMs are trained on a large number of words and have an intractable number of parameters, we make use of the Anchors approach to interpret the features learned by the model. This step is key for 1) making sure that the model does not pick up on spurious correlations from the huge dataset it has been trained on and 2) helping the medical practitioner understand why the model came to that particular decision. A key feature that makes anchors a suitable tool is that it is model-agnostic \citep{ribeiro2018anchors}. In our case, we took the predictions from the model and fed them into the anchors method. The default parameters in the implementation of the tool were used to interpret the model predictions. The anchor words extracted from the sentences of each cohort were then analyzed in the subsequent steps. In order to find the anchor words/tokens, the method perturbs each sentence from the data to see which words sufficiently explain the model prediction. This is done by assigning a probability value to the word/token and only words above a manually set probability threshold are considered as "anchor" words. In PainPoints we set this threshold to 0.95, which is the default. The perturbations are done by randomly replacing words in the predicted sentence by part-of-speech tags \citep{pennington2014glove} with a similar word in the embedding space. The method then outputs words which are identified as anchors. The idea is that if either of the anchor words are not present, then it is not sufficient to explain why the sentence belongs to that class.
\begin{figure}[ht]
\centering
\includegraphics[width=.45\textwidth]{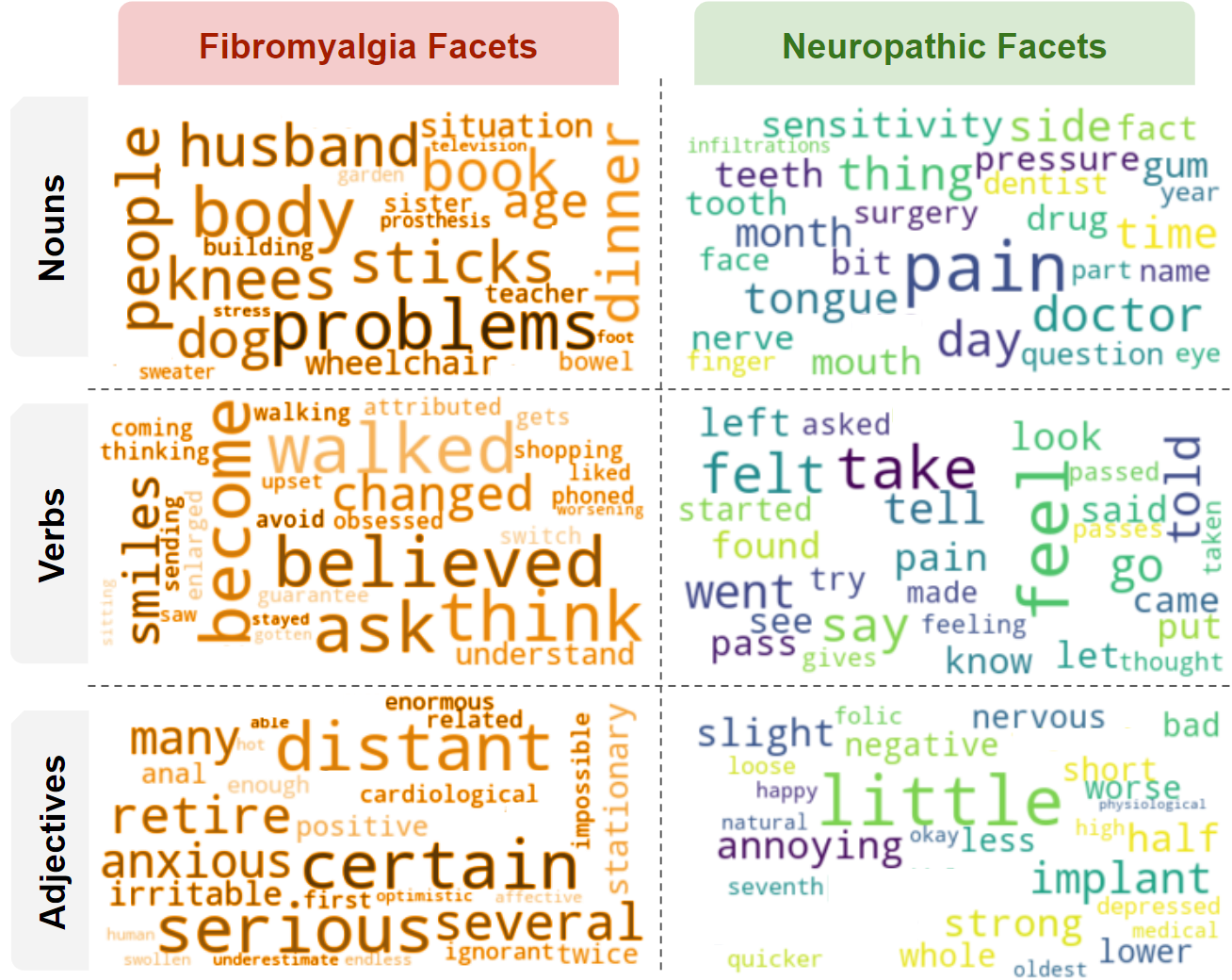}
\caption{Wordclouds of the top 50 facets obtained via fine-tuning BERT.}
\label{fig:anchors}
\end{figure}\\
\looseness=-1
\textbf{Expert-Collaborative Text Summaries:} While anchors extracted from the two cohort can be really useful to interpret the results, the main goal of PainPoints as a tool is to enable the medical practitioner to comprehend the results and use the insights in practice. Interview texts can be very long, depending on the duration of the patient interview (typically more than 3 pages of text). By harnessing the text summarization capabilities of LLMs, PainPoints enables generating interview summary notes. While the summaries are much easier to read and interpret, they run the risk of missing important points/ideas from the interview depending on the compression ratio one chooses to summarize the text with. This enables the practitioner to interactively discard useless aspects of the text. By leveraging the anchors derived from the previous step, PainPoints develops the idea of \textit{facets}. Facets are nothing but a union of all anchor words derived from the subjects in that particular cohort (FM or NP). Naturally, facets can be specific to an individual subject or can be common across subjects in that cohort. When intervening on a particular patient, the medical practitioner is given all facets present in the text document. Then by choosing which facets the practitioner wants to include in the summary, the sentences from the text are filtered and fed into the text summarization model. If the summarization ratio is too small, then the practitioner is shown what facets are missing and what sentences corresponded to those facets in the main document. This can be done interactively where the practitioner can add or remove facets chosen in the previous iteration. This re-summarization of the interview text allows for subjective selection of text which may have been lost in the process otherwise. It also helps the practitioner get a better intuition of what the model has learnt and implicitly incorporates the features considered important by the classifier in the first step. This feedback loop essentially enables an \textit{expert-in-the-loop} system that can be used for downstream tasks such as diagnosis, medication and treatment planning. \\
\looseness=-1
\textbf{Facet Coverage:} Quantifying the quality of the generated summary is of key importance in the proposed framework, because the expert interacting with the system needs a an estimate of what facets of the text were retained/lost in the process of summarization. Metrics proposed in the past that measure the quality of the summary such as BLEU scores compute the overlap between the words (n-gram) in the summary and the text it was generated from. However, these metrics do not reflect the ideas retained or lost in the summarization process, since they increase in proportion with the size of the generated summary. To overcome this shortcoming, we develop a new metric, Facet Coverage (FaCov), which aims to capture the common facets between the summary and original text. Instead of measuring the overlap between all the words on the original and generated text, FaCov measures the overlap between the facets in the summary and generated text. These facets are the same ones that were obtained from the model predictions in steps 1 and 2 of PainPoints. Let $X = {x_1, x_2, \dots x_n}$ be a set of all facets ($x_i$) in the original text and let $Y = {y_1, y_2, \dots y_n}$ be the facets ($y_i$) present in the generated summary. These facets can be generated by running the same anchors interpretability model on the summary. Additionally, if the medical practitioner wants to add extra facets based on prior knowledge, those can also be incorporated. Let the expert facets ($e_i$) be  $E = {e_1, e_2, \dots e_n}$. If the elements in both sets can be represented as $Z = (X \cap Y) \cup E$ based in the summary, then the FaCov score can be computed as $\frac{|Z|}{|X \cup E|}$ where $|.|$ implies cardinality of the set.
%$n(Z)$. 
\begin{figure}[ht]
\centering
\includegraphics[width=.45\textwidth]{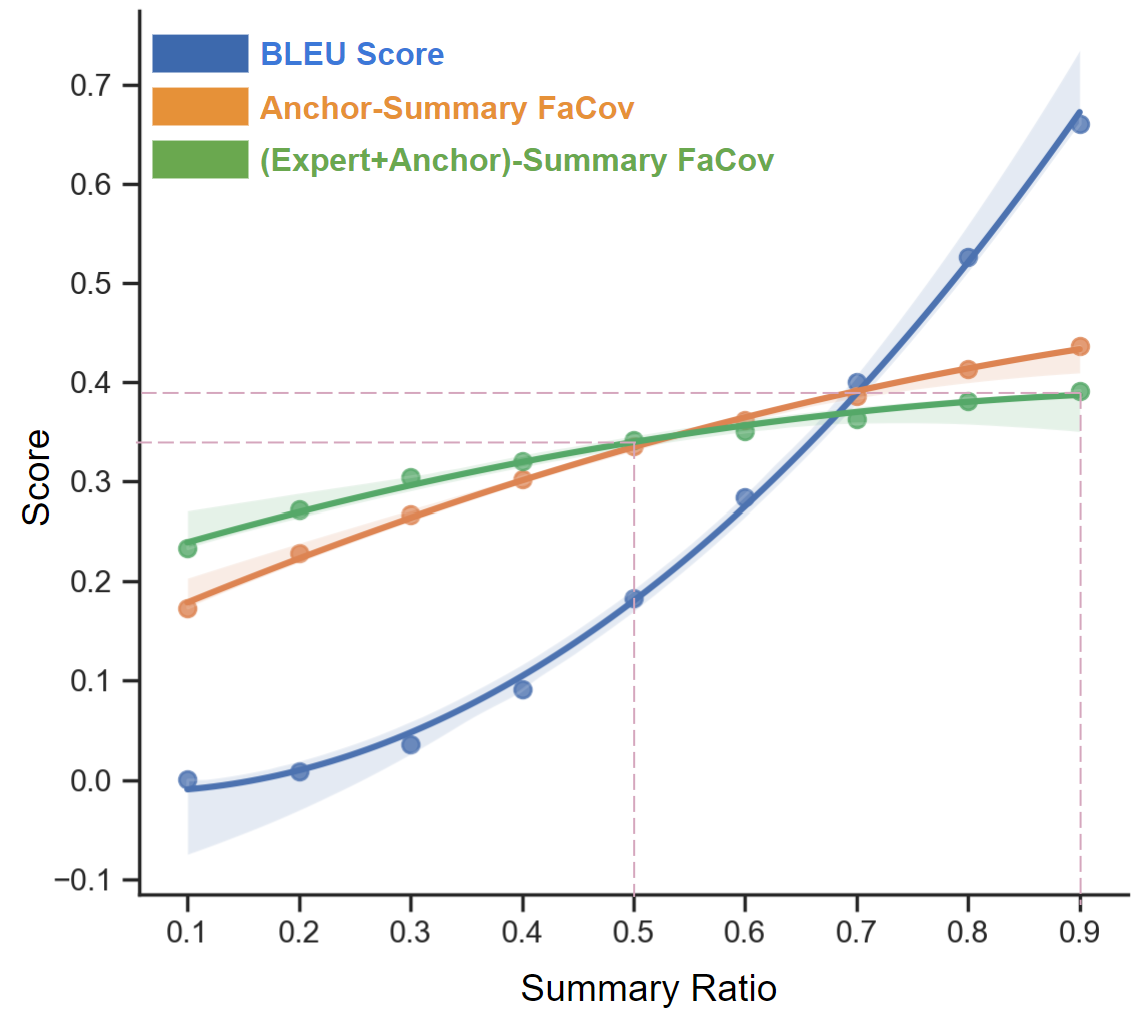}
\caption{Mean scores across all FM patients at different summarization ratios. Compares FaCov with and without expert facets against BLEU.}
\label{fig:score_comp}
\end{figure}

\begin{figure*}[ht]
\centering
\includegraphics[width=1\textwidth]{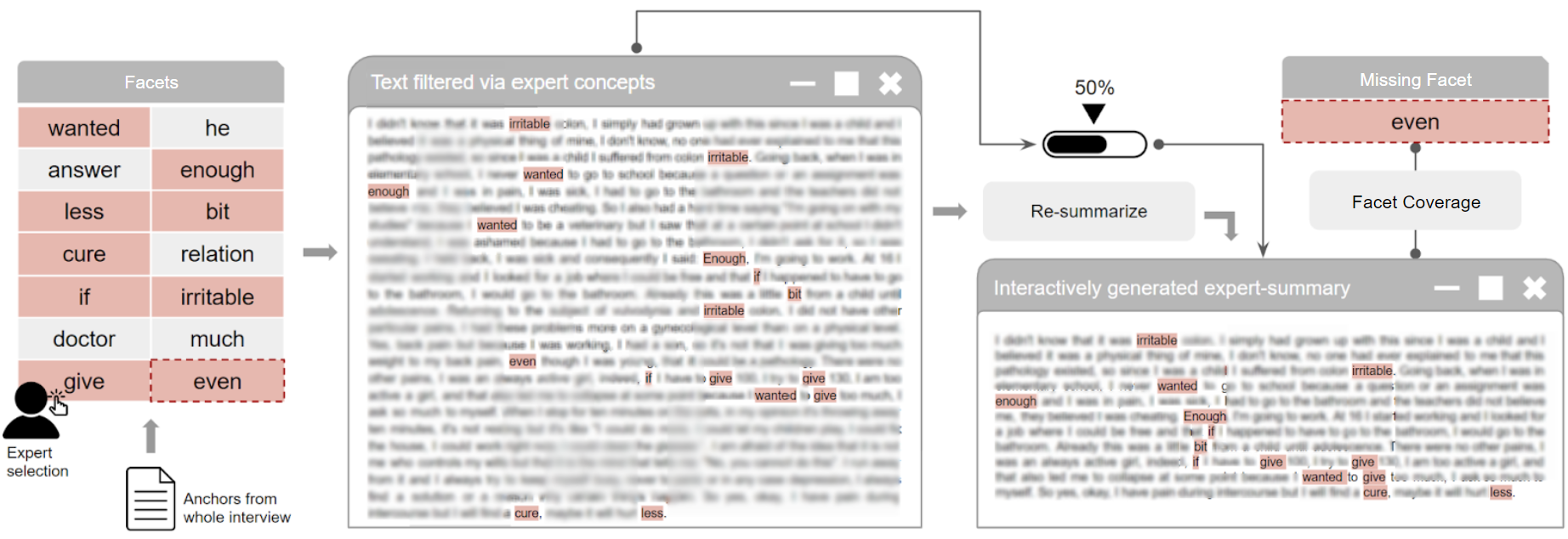}
\caption{Outlines the interactive flow of PainPoints where the medical practitioner can chose the facets to filter the raw documents and accordingly re-summarize the text. FaCov also reveals the missing facets from the summary.}
\label{fig:interactive}
\end{figure*}
\section{Results and Discussion}
\looseness=-1
To compare the performance of the classifiers, we compute the AUC scores for each of them. We use a shuffled 4-fold cross-validation and report the mean AUC in Table.~\ref{tab:auc_table}. The ROC curve for each classifier have been shown in Fig.~\ref{fig:rocs} and variance is estimated via the cross-validation scheme. On comparison, we see that the fine-tuned BERT significantly out-performs the other classifiers which are trained using the BERT embeddings by yielding an AUC of 0.83. While AdaBoost, Gaussian process and MLP achieve similar AUC scores (0.59), the RBF-kernel based SVM performs slightly better with an AUC of 0.62. This shows that by further fine-tuning BERT on the pain dataset, the pre-trained model is constrained to the text features prevalent in the patients. The fine-tuned model can find a better decision boundary between the two cohorts of NP and FM patients accurately. In Fig.~\ref{fig:ss_anchors} we show the different anchors picked up by the model via two randomly chosen sentences from both the NP and FM cohort. The anchors picked up by the model in the FM sentence are \textit{massage}, \textit{night} and \textit{excruciating}, whereas, the anchors in the NP sentence are \textit{tried}, \textit{everything}, \textit{pull}, \textit{feel} and \textit{well}. We segregated the anchors/ facets by their part-of-speech tags corresponding to nouns, verbs and adjectives for post hoc analysis. One can note that the words that FM patients use to describe their pain are very different from the NP patients. As explained in Sec.~1, this is expected as the manifestation of both the sub-types of chronic pain are very different from one another. Notably, we see nouns such as \textit{bowel} and \textit{dinner} in the FM cohort which is expected on account of the IBS diagnosis common to the cohort. Apart from these, the anchors in the FM cohort text appear to be related to lifestyle, social, and psychological experience such as \textit{situation}, \textit{problems}, \textit{stress}, and \textit{people}. On the other hand anchors of NP patients have a physical quality, like \textit{surgery}, \textit{teeth}, \textit{nerve}, and \textit{eye}. This result also abides by the differences in the symptom characteristics between the two cohorts: NP patients typically have physical nerve damage due to which they can pin-point their physical struggles. FM patients, however, experience a more widespread pain with sometimes co-occurring with psycho-social symptoms. Similarly, in the case of the verbs, FM patients tend to explain activities with words that represent psycho-social struggles such as \textit{thinking}, \textit{believed}, and \textit{upset}, and NP patients used more physical activity-related words like \textit{pain}, \textit{feel}, \textit{jump}, \textit{look}, and \textit{put}.  In Fig.~\ref{fig:interactive}, we show an outline of how the end user can interact with the facets and re-summarize the text. For instance, we show the selected words highlighted in red are used to filter the original text. Next, a summarization ratio is chosen to further compress the text and obtain summary notes of the interview. The facets selected in the original text are also highlighted in the summary. Using FaCov, we also indicate the missing facet from the summary (e.g. even). A key advantage of using FaCov is that it does not grow as rapidly as the BLEU score depicted in Fig.~\ref{fig:score_comp} (see dashed lines). We show this by computing FaCov and BLEU on all text documents of FM patients and plotting the mean score for each summarization ratio ($ [0.1, \dots, 0.9]$). The expert facets used to perform this group-level analysis are derived from expert questions which are detailed in the appendix. %The expert questions are tokenized and FaCov is computed using the common facets between the summary facets.
\section{Conclusion}
\looseness=-1
In this paper we proposed a novel human-AI collaborative tool, PainPoints, which learns to accurately identify sub-types of chronic pain (here, fibromyalgia and neuropathic pain) using fine-tuned large language models. PainPoints introduces a two-stage interpretability approach where the 1) anchor words in the text, which influence the classifier, are accurately detected and 2) the medical practitioner is allowed to interactively generate clinically usable summaries of the interview using these anchor words. In \textit{future work}, we will use prompt-engineering \cite{liu2021pre} to allow the practitioner to further tune facet-based-summaries.

% % add here
% While LLMs have shown remarkable accuracy on a variety of tasks and benchmarks, it has been shown that the accuracy of the model can further be improved by changing the way one interacts/draws inferences from the model. This has led to a new area of research known as \textit{prompting}. Prompting takes advantage of the fact that most models are trained using the MLM approach. Using this, one can use discrete or soft prompts (words) to better extract information from the model \citep{liu2021pre}. A prompt can be seen as a piece of text which is added to the input so that the model predictions can be better constrained to obtain a better result.

\bibliography{pmlr-sample}

\begin{thebibliography}{47}
\providecommand{\natexlab}[1]{#1}
\providecommand{\url}[1]{\texttt{#1}}
\expandafter\ifx\csname urlstyle\endcsname\relax
  \providecommand{\doi}[1]{doi: #1}\else
  \providecommand{\doi}{doi: \begingroup \urlstyle{rm}\Url}\fi

\bibitem[Berger and Baria(2022)]{berger2022}
S.~E. Berger and A.T. Baria.
\newblock Assessing pain research: a narrative review of emerging pain methods,
  their technosocial implications, and opportunities for multidisciplinary
  approaches, 2022.
\newblock \emph{Frontiers in Pain Research}, page~82, 2022.

\bibitem[Berger et~al.(2021)Berger, Branco, Vachon-Presseau, Abdullah, Cecchi,
  and Apkarian]{berger2021}
S.~E. Berger, P.~Branco, E.~Vachon-Presseau, T.~B. Abdullah, G.~Cecchi, and
  A.~V Apkarian.
\newblock Quantitative language features identify placebo responders in chronic
  back pain, 2021.
\newblock \emph{PAIN}, 162\penalty0 (6):\penalty0 1001, 2021.

\bibitem[Bonafede et~al.(1995)Bonafede, Downey, and
  Bennett]{bonafede1995association}
RP~Bonafede, DC~Downey, and RM~Bennett.
\newblock An association of fibromyalgia with primary sj{\"o}gren's syndrome: a
  prospective study of 72 patients.
\newblock \emph{The Journal of Rheumatology}, 22\penalty0 (1):\penalty0
  133--136, 1995.

\bibitem[Breivik et~al.(2008)Breivik, Borchgrevink, Allen, Rosseland,
  Romundstad, Breivik~Hals, Kvarstein, and Stubhaug]{breivik2008assessment}
Harald Breivik, Petter-Christian Borchgrevink, Sara-Maria Allen, Leiv-Arne
  Rosseland, Luis Romundstad, EK~Breivik~Hals, G~Kvarstein, and A~Stubhaug.
\newblock Assessment of pain.
\newblock \emph{BJA: British Journal of Anaesthesia}, 101\penalty0
  (1):\penalty0 17--24, 2008.

\bibitem[Brown et~al.(2020)Brown, Mann, Ryder, Subbiah, Kaplan, Dhariwal,
  Neelakantan, Shyam, Sastry, Askell, et~al.]{brown2020language}
Tom Brown, Benjamin Mann, Nick Ryder, Melanie Subbiah, Jared~D Kaplan, Prafulla
  Dhariwal, Arvind Neelakantan, Pranav Shyam, Girish Sastry, Amanda Askell,
  et~al.
\newblock Language models are few-shot learners.
\newblock \emph{Advances in neural information processing systems},
  33:\penalty0 1877--1901, 2020.

\bibitem[Buskila et~al.(1993)Buskila, Press, Gedalia, Klein, Neumann, Boehm,
  and Sukenik]{buskila1993assessment}
D~Buskila, J~Press, A~Gedalia, M~Klein, L~Neumann, R~Boehm, and S~Sukenik.
\newblock Assessment of nonarticular tenderness and prevalence of fibromyalgia
  in children.
\newblock \emph{The Journal of rheumatology}, 20\penalty0 (2):\penalty0
  368--370, 1993.

\bibitem[Colloca et~al.(2017)Colloca, Ludman, Bouhassira, Baron, Dickenson,
  Yarnitsky, Freeman, Truini, Attal, Finnerup, et~al.]{colloca2017neuropathic}
Luana Colloca, Taylor Ludman, Didier Bouhassira, Ralf Baron, Anthony~H
  Dickenson, David Yarnitsky, Roy Freeman, Andrea Truini, Nadine Attal, Nanna~B
  Finnerup, et~al.
\newblock Neuropathic pain.
\newblock \emph{Nature reviews Disease primers}, 3\penalty0 (1):\penalty0
  1--19, 2017.

\bibitem[Cortes and Vapnik(1995)]{cortes1995support}
Corinna Cortes and Vladimir Vapnik.
\newblock Support-vector networks.
\newblock \emph{Machine learning}, 20\penalty0 (3):\penalty0 273--297, 1995.

\bibitem[Dahlhamer et~al.(2016)Dahlhamer, Lucas, Zelaya, Nahin, Mackey, DeBar,
  Kerns, Von~Korff, Porter, and Helmick]{dahlhamer2018prevalence}
James Dahlhamer, Jacqueline Lucas, Carla Zelaya, Richard Nahin, Sean Mackey,
  Lynn DeBar, Robert Kerns, Michael Von~Korff, Linda Porter, and Charles
  Helmick.
\newblock Prevalence of chronic pain and high-impact chronic pain among
  adults—united states, 2016.
\newblock \emph{Morbidity and Mortality Weekly Report}, 67\penalty0
  (36):\penalty0 1692--1704, 2016.

\bibitem[Devlin et~al.(2018)Devlin, Chang, Lee, and Toutanova]{devlin2018bert}
Jacob Devlin, Ming-Wei Chang, Kenton Lee, and Kristina Toutanova.
\newblock Bert: Pre-training of deep bidirectional transformers for language
  understanding.
\newblock \emph{arXiv preprint arXiv:1810.04805}, 2018.

\bibitem[Dhurandhar et~al.(2018)Dhurandhar, Shanmugam, Luss, and
  Olsen]{dhurandhar2018improving}
Amit Dhurandhar, Karthikeyan Shanmugam, Ronny Luss, and Peder~A Olsen.
\newblock Improving simple models with confidence profiles.
\newblock \emph{Advances in Neural Information Processing Systems}, 31, 2018.

\bibitem[Dhurandhar et~al.(2020)Dhurandhar, Shanmugam, and
  Luss]{dhurandhar2020enhancing}
Amit Dhurandhar, Karthikeyan Shanmugam, and Ronny Luss.
\newblock Enhancing simple models by exploiting what they already know.
\newblock In \emph{International Conference on Machine Learning}, pages
  2525--2534. PMLR, 2020.

\bibitem[Ehde et~al.(2005)Ehde, Osborne, and Jensen]{ehde2005chronic}
Dawn~M Ehde, Travis~L Osborne, and Mark~P Jensen.
\newblock Chronic pain in persons with multiple sclerosis.
\newblock \emph{Physical Medicine and Rehabilitation Clinics}, 16\penalty0
  (2):\penalty0 503--512, 2005.

\bibitem[Frohman and Wingerchuk(2010)]{frohman2010transverse}
Elliot~M Frohman and Dean~M Wingerchuk.
\newblock Transverse myelitis.
\newblock \emph{New England Journal of Medicine}, 363\penalty0 (6):\penalty0
  564--572, 2010.

\bibitem[H{\"a}user et~al.(2015)H{\"a}user, Ablin, Fitzcharles, Littlejohn,
  Luciano, Usui, and Walitt]{hauser2015fibromyalgia}
Winfried H{\"a}user, Jacob Ablin, Mary-Ann Fitzcharles, Geoffrey Littlejohn,
  Juan~V Luciano, Chie Usui, and Brian Walitt.
\newblock Fibromyalgia.
\newblock \emph{Nature reviews Disease primers}, 1\penalty0 (1):\penalty0
  1--16, 2015.

\bibitem[Hinton et~al.(2015)Hinton, Vinyals, Dean,
  et~al.]{hinton2015distilling}
Geoffrey Hinton, Oriol Vinyals, Jeff Dean, et~al.
\newblock Distilling the knowledge in a neural network.
\newblock \emph{arXiv preprint arXiv:1503.02531}, 2\penalty0 (7), 2015.

\bibitem[Howard and Ruder(2018)]{howard2018universal}
Jeremy Howard and Sebastian Ruder.
\newblock Universal language model fine-tuning for text classification.
\newblock \emph{arXiv preprint arXiv:1801.06146}, 2018.

\bibitem[Kaplan et~al.(2020)Kaplan, McCandlish, Henighan, Brown, Chess, Child,
  Gray, Radford, Wu, and Amodei]{kaplan2020scaling}
Jared Kaplan, Sam McCandlish, Tom Henighan, Tom~B Brown, Benjamin Chess, Rewon
  Child, Scott Gray, Alec Radford, Jeffrey Wu, and Dario Amodei.
\newblock Scaling laws for neural language models.
\newblock \emph{arXiv preprint arXiv:2001.08361}, 2020.

\bibitem[LeCun et~al.(2015)LeCun, Bengio, and Hinton]{lecun2015deep}
Yann LeCun, Yoshua Bengio, and Geoffrey Hinton.
\newblock Deep learning.
\newblock \emph{nature}, 521\penalty0 (7553):\penalty0 436--444, 2015.

\bibitem[Lee et~al.(2022)Lee, Liang, and Yang]{lee2022coauthor}
Mina Lee, Percy Liang, and Qian Yang.
\newblock Coauthor: Designing a human-ai collaborative writing dataset for
  exploring language model capabilities.
\newblock In \emph{CHI Conference on Human Factors in Computing Systems}, pages
  1--19, 2022.

\bibitem[Lin(2004)]{lin2004rouge}
Chin-Yew Lin.
\newblock Rouge: A package for automatic evaluation of summaries.
\newblock In \emph{Text summarization branches out}, pages 74--81, 2004.

\bibitem[Liu et~al.(2021)Liu, Yuan, Fu, Jiang, Hayashi, and Neubig]{liu2021pre}
Pengfei Liu, Weizhe Yuan, Jinlan Fu, Zhengbao Jiang, Hiroaki Hayashi, and
  Graham Neubig.
\newblock Pre-train, prompt, and predict: A systematic survey of prompting
  methods in natural language processing.
\newblock \emph{arXiv preprint arXiv:2107.13586}, 2021.

\bibitem[Liu et~al.(2019)Liu, Ott, Goyal, Du, Joshi, Chen, Levy, Lewis,
  Zettlemoyer, and Stoyanov]{liu2019roberta}
Yinhan Liu, Myle Ott, Naman Goyal, Jingfei Du, Mandar Joshi, Danqi Chen, Omer
  Levy, Mike Lewis, Luke Zettlemoyer, and Veselin Stoyanov.
\newblock Roberta: A robustly optimized bert pretraining approach.
\newblock \emph{arXiv preprint arXiv:1907.11692}, 2019.

\bibitem[Loper and Bird(2002)]{loper2002nltk}
Edward Loper and Steven Bird.
\newblock Nltk: The natural language toolkit.
\newblock \emph{arXiv preprint cs/0205028}, 2002.

\bibitem[Lundberg and Lee(2017)]{lundberg2017unified}
Scott~M Lundberg and Su-In Lee.
\newblock A unified approach to interpreting model predictions.
\newblock \emph{Advances in neural information processing systems}, 30, 2017.

\bibitem[Luss and Dhurandhar(2021)]{luss2021towards}
Ronny Luss and Amit Dhurandhar.
\newblock Towards better model understanding with path-sufficient explanations.
\newblock \emph{arXiv preprint arXiv:2109.06181}, 2021.

\bibitem[Martinez-Lavin(2007)]{martinez2007biology}
Manuel Martinez-Lavin.
\newblock Biology and therapy of fibromyalgia. stress, the stress response
  system, and fibromyalgia.
\newblock \emph{Arthritis research \& therapy}, 9\penalty0 (4):\penalty0 1--7,
  2007.

\bibitem[Nickisch and Rasmussen(2008)]{nickisch2008approximations}
Hannes Nickisch and Carl~Edward Rasmussen.
\newblock Approximations for binary gaussian process classification.
\newblock \emph{Journal of Machine Learning Research}, 9\penalty0
  (Oct):\penalty0 2035--2078, 2008.

\bibitem[Papineni et~al.(2002)Papineni, Roukos, Ward, and
  Zhu]{papineni2002bleu}
Kishore Papineni, Salim Roukos, Todd Ward, and Wei-Jing Zhu.
\newblock Bleu: a method for automatic evaluation of machine translation.
\newblock In \emph{Proceedings of the 40th annual meeting of the Association
  for Computational Linguistics}, pages 311--318, 2002.

\bibitem[Pennington et~al.(2014)Pennington, Socher, and
  Manning]{pennington2014glove}
Jeffrey Pennington, Richard Socher, and Christopher~D Manning.
\newblock Glove: Global vectors for word representation.
\newblock In \emph{Proceedings of the 2014 conference on empirical methods in
  natural language processing (EMNLP)}, pages 1532--1543, 2014.

\bibitem[Peters et~al.(2018)Peters, Neumann, Zettlemoyer, and
  Yih]{peters2018dissecting}
Matthew~E Peters, Mark Neumann, Luke Zettlemoyer, and Wen-tau Yih.
\newblock Dissecting contextual word embeddings: Architecture and
  representation.
\newblock \emph{arXiv preprint arXiv:1808.08949}, 2018.

\bibitem[Pisner and Schnyer(2020)]{pisner2020support}
Derek~A Pisner and David~M Schnyer.
\newblock Support vector machine.
\newblock In \emph{Machine learning}, pages 101--121. Elsevier, 2020.

\bibitem[Radford et~al.(2018)Radford, Narasimhan, Salimans, Sutskever,
  et~al.]{radford2018improving}
Alec Radford, Karthik Narasimhan, Tim Salimans, Ilya Sutskever, et~al.
\newblock Improving language understanding by generative pre-training.
\newblock \emph{OpenAI}, 2018.

\bibitem[Radford et~al.(2019)Radford, Wu, Child, Luan, Amodei, Sutskever,
  et~al.]{radford2019language}
Alec Radford, Jeffrey Wu, Rewon Child, David Luan, Dario Amodei, Ilya
  Sutskever, et~al.
\newblock Language models are unsupervised multitask learners.
\newblock \emph{OpenAI blog}, 1\penalty0 (8):\penalty0 9, 2019.

\bibitem[Ribeiro et~al.(2016)Ribeiro, Singh, and Guestrin]{ribeiro2016should}
Marco~Tulio Ribeiro, Sameer Singh, and Carlos Guestrin.
\newblock " why should i trust you?" explaining the predictions of any
  classifier.
\newblock In \emph{Proceedings of the 22nd ACM SIGKDD international conference
  on knowledge discovery and data mining}, pages 1135--1144, 2016.

\bibitem[Ribeiro et~al.(2018)Ribeiro, Singh, and Guestrin]{ribeiro2018anchors}
Marco~Tulio Ribeiro, Sameer Singh, and Carlos Guestrin.
\newblock Anchors: High-precision model-agnostic explanations.
\newblock In \emph{Proceedings of the AAAI conference on artificial
  intelligence}, volume~32, 2018.

\bibitem[Scandola et~al.(2021)Scandola, Pietroni, Landuzzi, Polati, Schweiger,
  and Moro]{scandola2021bodily}
Michele Scandola, Giorgia Pietroni, Gabriella Landuzzi, Enrico Polati, Vittorio
  Schweiger, and Valentina Moro.
\newblock Bodily illusions and motor imagery in fibromyalgia.
\newblock \emph{Frontiers in human neuroscience}, 15, 2021.

\bibitem[Schapire(2013)]{schapire2013explaining}
Robert~E Schapire.
\newblock Explaining adaboost.
\newblock In \emph{Empirical inference}, pages 37--52. Springer, 2013.

\bibitem[Schweiger et~al.(2022)Schweiger, Secchettin, Perini, Martini,
  Donadello, Gottin, Del~Balzo, Varrassi, and Polati]{schweiger2022quality}
Vittorio Schweiger, Erica Secchettin, Giovanni Perini, Alvise Martini, Katia
  Donadello, Leonardo Gottin, Giovanna Del~Balzo, Giustino Varrassi, and Enrico
  Polati.
\newblock Quality of life and psychological assessment in patients with
  fibromyalgia syndrome during covid-19 pandemic in italy: prospective
  observational study.
\newblock \emph{Signa Vitae}, 2022.

\bibitem[Sperber et~al.(1999)Sperber, Atzmon, Neumann, Weisberg, Shalit,
  Abu-Shakrah, Fich, and Buskila]{sperber1999fibromyalgia}
AD~Sperber, Y~Atzmon, L~Neumann, I~Weisberg, Y~Shalit, M~Abu-Shakrah, A~Fich,
  and D~Buskila.
\newblock Fibromyalgia in the irritable bowel syndrome: studies of prevalence
  and clinical implications.
\newblock \emph{The American journal of gastroenterology}, 94\penalty0
  (12):\penalty0 3541--3546, 1999.

\bibitem[Sun et~al.(2019)Sun, Wang, Li, Feng, Chen, Zhang, Tian, Zhu, Tian, and
  Wu]{sun2019ernie}
Yu~Sun, Shuohuan Wang, Yukun Li, Shikun Feng, Xuyi Chen, Han Zhang, Xin Tian,
  Danxiang Zhu, Hao Tian, and Hua Wu.
\newblock Ernie: Enhanced representation through knowledge integration.
\newblock \emph{arXiv preprint arXiv:1904.09223}, 2019.

\bibitem[Vaswani et~al.(2017)Vaswani, Shazeer, Parmar, Uszkoreit, Jones, Gomez,
  Kaiser, and Polosukhin]{vaswani2017attention}
Ashish Vaswani, Noam Shazeer, Niki Parmar, Jakob Uszkoreit, Llion Jones,
  Aidan~N Gomez, {\L}ukasz Kaiser, and Illia Polosukhin.
\newblock Attention is all you need.
\newblock \emph{Advances in neural information processing systems}, 30, 2017.

\bibitem[Wallace et~al.(2019)Wallace, Tuyls, Wang, Subramanian, Gardner, and
  Singh]{wallace2019allennlp}
Eric Wallace, Jens Tuyls, Junlin Wang, Sanjay Subramanian, Matt Gardner, and
  Sameer Singh.
\newblock Allennlp interpret: A framework for explaining predictions of nlp
  models.
\newblock \emph{arXiv preprint arXiv:1909.09251}, 2019.

\bibitem[Wang et~al.(2017)Wang, Sung, Men, Wang, Lin, and
  Kao]{wang2017bidirectional}
Jia-Chi Wang, Fung-Chang Sung, Mauranda Men, Kevin~A Wang, Cheng-Li Lin, and
  Chia-Hung Kao.
\newblock Bidirectional association between fibromyalgia and gastroesophageal
  reflux disease: two population-based retrospective cohort analysis.
\newblock \emph{Pain}, 158\penalty0 (10):\penalty0 1971--1978, 2017.

\bibitem[Wolf et~al.(2019)Wolf, Debut, Sanh, Chaumond, Delangue, Moi, Cistac,
  Rault, Louf, Funtowicz, et~al.]{wolf2019huggingface}
Thomas Wolf, Lysandre Debut, Victor Sanh, Julien Chaumond, Clement Delangue,
  Anthony Moi, Pierric Cistac, Tim Rault, R{\'e}mi Louf, Morgan Funtowicz,
  et~al.
\newblock Huggingface's transformers: State-of-the-art natural language
  processing.
\newblock \emph{arXiv preprint arXiv:1910.03771}, 2019.

\bibitem[Yang et~al.(2019)Yang, Dai, Yang, Carbonell, Salakhutdinov, and
  Le]{yang2019xlnet}
Zhilin Yang, Zihang Dai, Yiming Yang, Jaime Carbonell, Russ~R Salakhutdinov,
  and Quoc~V Le.
\newblock Xlnet: Generalized autoregressive pretraining for language
  understanding.
\newblock \emph{Advances in neural information processing systems}, 32, 2019.

\bibitem[Zelaya et~al.(2020)Zelaya, Dahlhamer, Lucas, and
  Connor]{zelaya2020chronic}
Carla~E Zelaya, James~M Dahlhamer, Jacqueline~W Lucas, and Eric~M Connor.
\newblock Chronic pain and high-impact chronic pain among us adults, 2019.
\newblock \emph{National Center for Health Statistics Data Brief No. 390},
  2020.

\end{thebibliography}

\appendix
\section{Additional Metrics}
\begin{figure*}[ht]
\centering
\includegraphics[width=1\textwidth]{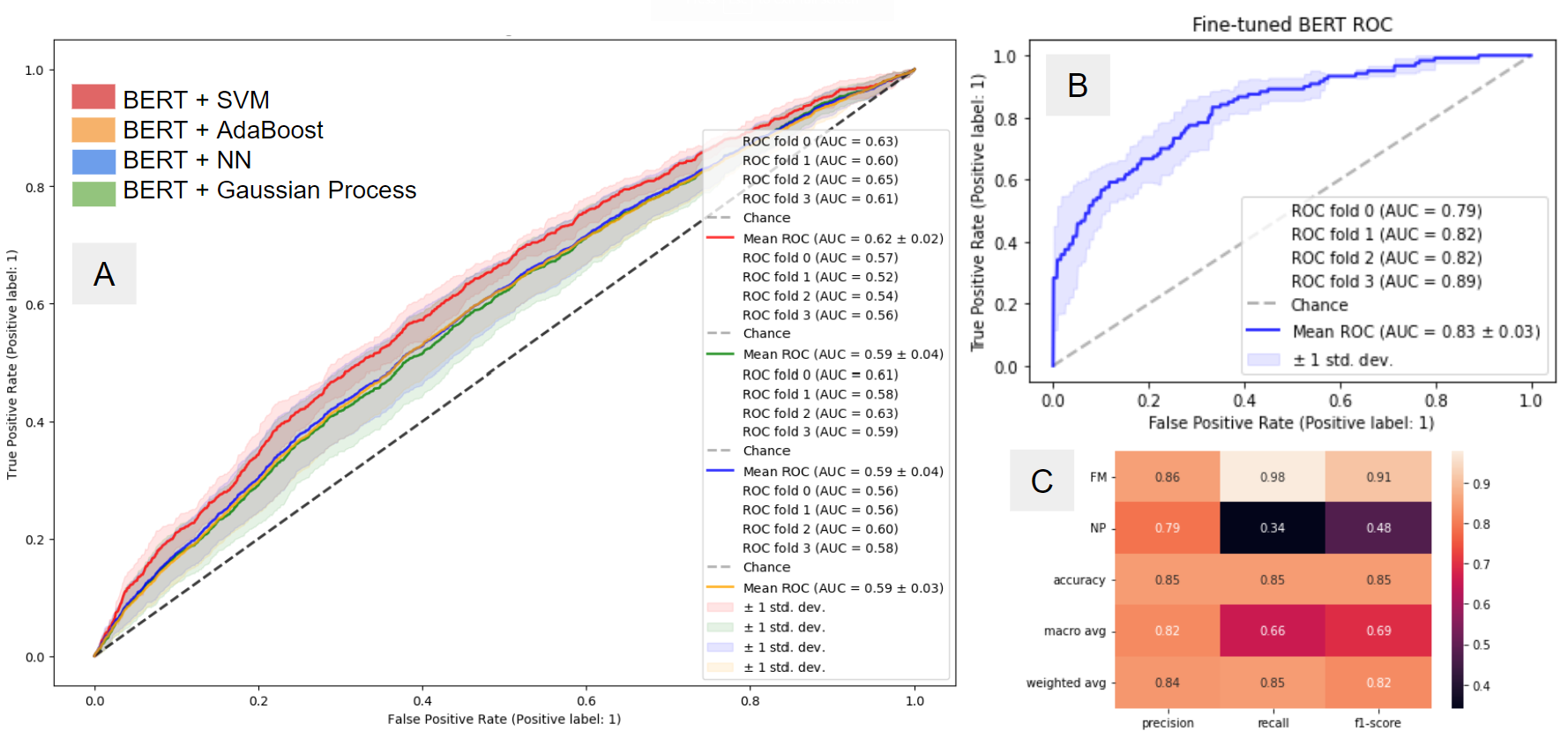}
\caption{Shows Mean AUC scores along with 1 standard deviation for each predictive model ([A] Shows BERT Embeddings + different classifiers, [B] ). Also tabulates the different accuracy metrics in [C].}
\label{fig:extra}
\end{figure*}

In the Fig.~\ref{fig:extra} we show the AUC score obtained per-fold from each predictive model. 1 standard deviation along with the mean score is also reported. Additionally, metrics such as precision, recall, accuracy for both cohorts are reported.

\section{Compute Resources}
All experiments pertaining to the steps within PainPoints were performed on a standard Windows i7 laptop with 16GB RAM. The BERT model was Fine-tuned on a single NVIDIA P100 GPU. The fine-tuning takes less than 5 minutes while the computation time of anchors and FaCov scores is negligible on a standard i7 CPU. Scikit-Learn was used to perform metric comparsions, cross-validation and to implement the SVM, GP, NN and AdaBoost algorithms.

\section{Data and Software}
The data used to perform the analysis has been included in the anonymous link below. The BERT pre-trained model was used via the transformers package of HuggingFace \citep{wolf2019huggingface}. The implementation was done in PyTorch and Python. NLTK \citep{loper2002nltk} and SpaCy [{\url{https://github.com/explosion/spaCy}}] were used to perform pre-processing tasks such as tokenization and part-of-speech tagging. The dependencies are detailed in the README withing the same anonymous link below.

The data used in the paper and its derivatives are made available at the following anonymous link: \textcolor{green}{\url{https://figshare.com/s/5249898bf5514457372a}}. The code and instructions to reproduce the results have been added to the folder in a \textcolor{red}{README} file.

\section{Expert Facets}
Expert Facets in the paper are extracted from the following questions:

"Do you experience little interest or pleasure in doing things?"\\
"Are you feeling down, depressed, or hopeless?"\\
"Do you have trouble falling or staying asleep?"\\
"Do you feel that you are sleeping too much?"\\
"Do you feel tired or having little energy?"\\
"Do you have poor appetite"\\
"Have you been overeating?"\\
"Do you feel bad about yourself - or that you are a failure or have let yourself or your family down?"\\
"Do you have trouble concentrating on things, such as reading the newspaper or watching television?"\\
"Are you moving or speaking so slowly that other people could have noticed?"\\
"Have you been so fidgety or restless that you have been moving around a lot more than usual?"\\
"Do you have thoughts that you would be better off dead, or of hurting yourself in some way?"
\end{document}